Draft: Finding structure in data using multivariate tree boosting.

Patrick J. Miller, Gitta H. Lubke, Daniel B. McArtor, C. S. Bergeman

University of Notre Dame

Author Note

Patrick J. Miller, Department of Psychology, University of Notre Dame; Gitta H. Lubke, Department of Psychology, University of Notre Dame; Daniel B. McArtor, Department of Psychology, University of Notre Dame; C.S. Bergeman, Department of Psychology, University of Notre Dame.

We are grateful for the comments received from reviewers, Scott E. Maxwell, Ian Campbell, and Justin Lunningham on earlier versions of this manuscript.

This research was based upon work supported by the National Science Foundation Graduate Research Fellowship Program under grant number 1313583. The second author is supported by NIDA R37 DA-018673. The fourth author is supported by a grant from the National Institute of Aging (1 R01 AG023571-A1-01). The computational work was done on clusters acquired through NSF MRI BCS-1229450.

Correspondence concerning this article should be addressed to Patrick J. Miller & Gitta H. Lubke, 110 Haggar Hall, University of Notre Dame, Notre Dame, IN 46656. Email: pmille13@nd.edu, glubke@nd.edu

This revision was submitted to Psychological Methods, January 15, 2016.

**Abstract**

Technology and collaboration enable dramatic increases in the size of psychological and psychiatric data collections, but finding structure in these large data sets with many collected variables is challenging. Decision tree ensembles like random forests (Strobl, Malley, & Tutz, 2009) are a useful tool for finding structure, but are difficult to interpret with multiple outcome variables which are often of interest in psychology. To find and interpret structure in data sets with multiple outcomes and many predictors (possibly exceeding the sample size), we introduce a multivariate extension to a decision tree ensemble method called Gradient Boosted Regression Trees (Friedman, 2001). Our extension, *multivariate tree boosting*, is a method for non-parametric regression that is useful for identifying important predictors, detecting predictors with non-linear effects and interactions without specification of such effects, and for identifying predictors that cause two or more outcome variables to covary. We provide the R package '*mvtboost*' to estimate, tune, and interpret the resulting model, which extends the implementation of univariate boosting in the R package '*gbm*' (Ridgeway et al., 2015) to continuous, multivariate outcomes. To illustrate the approach, we analyze predictors of psychological well-being (Ryff & Keyes, 1995). Simulations verify that our approach identifies predictors with non-linear effects and achieves high prediction accuracy, exceeding or matching the performance of (penalized) multivariate multiple regression and multivariate decision trees over a wide range of conditions.

*Keywords*: boosting, multivariate, decision trees, ensemble, non-parametric regression, model selection,

Finding structure in data using multivariate tree boosting.

Technology and collaboration enable dramatic increases in the size of psychological and psychiatric data collections, in terms of both the overall sample size and the number of variables collected. A major challenge is to develop and evaluate methods that can serve to leverage the information in these growing data sets to better understand human behavior. Big data can take many forms, such as social media data, audio and video recordings, web-site logs, genetic sequences, and medical records (Chen, Chiang, & Storey, 2012; Howe et al., 2008; Manovich, 2012). In psychology and psychiatry, big data often take the form of large surveys with hundreds of individual questionnaire items comprised of many outcomes and predictors. In this context, however, it can be difficult to know what types of models to consider or even which variables to include in the model. As a result, it is often the case that many possible models need to be explored to find the model that most adequately captures the structure in the observed data. Although parametric models like multivariate multiple regression and SEM can be used in this model selection, these methods make strong structural and distributional assumptions that limit exploration. To address this limitation, the current paper describes *multivariate tree boosting,* a machine learning alternative to comparing different parametric models that more easily allows discovery of important structural features in observed variables, such as non-linear or interaction effects, and the detection of predictors that affect only some of the outcome variables.

Finding structure in observed data even in the absence of strong theory is particularly important for enhancing construct and external validity and for examining possible validity threats (Shadish, Cook, & Campbell, 2002). For instance, in the context of psychological testing, it is important to discover grouping variables (such as age or sex) that influence particular items in a test, indicating differential item functioning (Holland & Wainer, 1993). For observational studies, it is important to identify predictors with non-linear effects or predictors that interact because presence of these effects makes the interpretation of main or linear effects misleading. In an experimental design, interactions between a treatment and other covariates can indicate limits of generalization (Shadish et al., 2002). More broadly, detecting and interpreting non-linear and predictor-specific effects can enhance the development of theory.

The usual approach for finding structure among many predictors and outcomes is to fit and compare a limited number of parametric models (Burnham & Anderson, 2002). These parametric models can involve latent variables (e.g. SEM, factor analysis) or only observed variables (e.g. canonical correlation analysis, multivariate multiple regression). But there are significant problems with using parametric models for data exploration, including exploration in big data sets. First, in addition to distributional assumptions, models often make the strong assumption that a system of linear equations can sufficiently capture the important structure in the observed data. This ignores non-linear effects and possible interactions, unless they are explicitly specified. Second, because the structure of the data is typically unknown, the number of models that must be included in a comparison for a thorough exploration can easily become unwieldy. Even if some non-linear or interaction effects are included in the model, it can be difficult to specify all of the potentially relevant direct effects, non-linear effects, and interactions in a parametric model *a priori*. In addition, it is impossible to estimate these effects simultaneously if the number of effects is larger than the sample size. Third, model selection is usually *ad hoc* rather than systematic, and is not guaranteed to capture all of the important structural features. Even automatic procedures like step-wise regression and best subsets analysis are known to be unable to identify a correct set of predictors of a given size, and capitalize on sampling error (Hocking & Leslie, 1967; Thompson, 1995). Finally, conducting inference after this type of model selection inflates Type I Error and leads to results that are difficult to replicate (Burnham & Anderson, 2002; Gelman & Loken, 2013).

An alternative to model selection using parametric models is to use a non-parametric approach like decision trees (Breiman, Friedman, Stone, & Olshen, 1984). Decision trees are powerful because they can approximate non-linear effects and interactions and handle missing data in the predictors. This is done without making parametric assumptions about the structure of the observed data. Decision trees are also easily interpretable in terms of decision rules, which are defined by the splitting variables. Each decision rule is in the form of a conditional effect (e.g., if $x_1 < c$ and $x_2 < d$ then…), which defines groups of observations with similar scores. Although decision trees are flexible and easy to interpret, the estimated structure varies considerably from sample to sample. Bagging (*b*ootstrap *agg*regation, Breiman, 1996) and random forests (Breiman, 2001), which includes bagging, improve on single decision trees by fitting many decision trees to bootstrap samples, forming an ensemble that is more robust against random sampling fluctuation (Strobl et al., 2009).

Decision trees and random forests have been successfully used in observational studies in psychology to identify predictors of mid- and later life stress (Scott, Jackson, & Bergeman, 2011; Scott, Whitehead, Bergeman, & Pitzer, 2013), well-being in later life (Wallace, Bergeman, & Maxwell, 2002) and caregiver stability (Proctor et al., 2011). Decision trees have been used clinically to predict suicide attempts in psychiatric patients (Mann et al., 2008), and in experimental designs to analyze the effects of competence on depressive symptoms in children (Seroczynski, Cole, & Maxwell, 1997).

There are already several ways to extend decision trees to multivariate outcomes, including many variants of multivariate decision trees (Brodley & Utgoff, 1995; Brown, Pittard, & Park, 1996; De'Ath, 2002; Dine, Larocque, & Bellavance, 2009; Franco-Arcega, Carrasco-Ochoa, Sánchez-Díaz, & Martínez-Trinidad, 2010; Hsiao & Shih, 2007; Struyf & Džeroski, 2006) and decision trees for longitudinal outcomes (Loh & Zheng, 2013; Segal, 1992; Sela & Simonoff, 2012). Decision tree methods can also be extended to multivariate outcomes by combining decision trees with parametric models, in which model parameters are allowed to differ in groups defined by split points on covariates (Zeileis, Hothorn, & Hornik, 2008). *Structural Equation Model Trees* (Brandmaier, von Oertzen, McArdle, & Lindenberger, 2013) are an example of this approach, in which parameters of a structural equation model are allowed to differ in each partition. SEM trees can be used to address possible sources of measurement non-invariance, to detect group differences on factors, or to detect group differences in trajectories for longitudinal data (Brandmaier et al., 2013).

A natural extension to multivariate decision trees is a multivariate decision forest (Hothorn, Hornik, & Zeileis, 2006; Segal & Xiao, 2011). Like decision ensembles for single response variables, the multivariate random forest provides critical improvements to predictive performance compared to multivariate decision trees by combining predictions from many decision trees. Another example of a multivariate decision forest is a SEM forest with a fully saturated model for the covariance matrix, with the only difference being that the SEM forest uses a maximum likelihood criterion for split evaluation (Brandmaier et al., 2013). However, one of the limitations to ensembles of multivariate trees is that they are difficult to interpret. In general, decision tree ensembles exchange interpretability for prediction performance.

To address the limitations of exploratory analyses with parametric models and the difficulty of interpreting multivariate tree ensembles, we propose a new approach for exploratory data analysis with multivariate outcomes called *multivariate tree boosting*. Multivariate tree boosting is an extension of boosting for univariate outcomes (Bühlmann & Hothorn, 2007; Bühlmann & Yu, 2003; Freund & Schapire, 1997; Friedman, 2001, 2002). It fits univariate trees to multivariate outcomes by selecting the tree that maximizes the covariance explained in the outcomes. Multivariate tree boosting addresses the problem of model selection with multivariate outcomes by smoothly approximating non-linear effects and interactions by additive models of trees without requiring specification of these effects *a priori*. The method is suitable for truly big-data scenarios in which the number of predictors is only limited by available memory and computation time. Its flexibility also makes it useful for exploratory analyses involving only a few variables. Our method differs from SEM trees/forests by not requiring a model to be specified for the outcome variables. It differs from multivariate decision forests (or saturated SEM forests) by allowing easier interpretation of non-linear effects of predictors on individual outcome variables.

Our approach specifically addresses the issue of interpretation of multivariate tree ensembles because interpretation of the model is critical for applications in psychology and other sciences. We describe in detail how to use the model to select important variables, visualize non-linear effects, and detect departures from additivity. Multivariate tree boosting also allows estimation of the covariance explained in pairs of outcomes by predictors, a novel interpretation we think will be relevant for exploratory analyses in psychology. Our R package called '*mvtboost*' makes it easy to fit, tune, and interpret a multivariate tree boosting model, and extends the implementation of univariate boosting in the R package '*gbm*' (Ridgeway et al., 2015) to multivariate outcomes.

Below, we introduce the approach by describing decision trees and univariate boosting, followed by introducing multivariate tree boosting. We demonstrate how to estimate, tune, and interpret the multivariate tree boosting model using functions in *'mvtboost'*. Specifically, we use multivariate tree boosting to identify predictors that contribute to specific aspects (sub-scales) of psychological well-being in aging adults. This example illustrates how multivariate tree boosting can be used to answer exploratory questions such as: Which predictors are important? What is the functional form of the effect of important predictors? How do predictors explain covariance in the outcomes? Finally, we use a simulation to evaluate the prediction error and prediction selection performance of our approach compared to other model based and exploratory approaches.

## Decision Trees

**Description**. Decision trees use a series of dichotomous splits on predictor variables to create groups of observations (nodes) that are maximally homogeneous with respect to the outcome variable. Nodes in a tree are created by finding both the predictor and the optimal split point on that predictor that result in maximally-increased homogeneity in the daughter nodes. For continuous outcome variables, this means choosing predictors and split points that minimize the sums of squared errors within each daughter node. This process of finding a splitting variable and split point is then repeated within each daughter node, and is called recursive partitioning (Strobl et al., 2009).

There are several different ways to understand decision trees, which emphasize different properties. Decision trees are often represented as tree diagrams (Figure 1A), which show the set of variables and split points used to form the nodes. These diagrams represent the decision rules used to identify groups that are similar with respect to the outcome variable. Trees can also be viewed as models of conditional effects: the predictor with the largest main effect is selected for the first split, and each subsequent split is an effect conditional on all previously selected predictors. Thus trees can capture interactions between predictors. From a geometric perspective, a decision tree is a piecewise function (Figure 1B). The decision rules or splits form regions or nodes (denoted $R_j$) in the predictor space, and the predicted values of a tree are the means within each of these regions. Algebraically, a tree can be represented by the piecewise function $T(X,\theta) = \sum_{j=1}^{J} \gamma_j I(X \in R_j)$ where $\theta$ contains the split points and predictors defining $J$ regions $R_j$, and $\gamma_j$ are the predictions in each region (Friedman, 2001). The indicator function, $I(X \in R_j)$ denotes which observations in $X$ fall into region $R_j$. Thus the tree, $T(X,\theta)$, is a piecewise approximation of the unknown but potentially complex and non-linear function $F(X)$ relating the outcome variable $y$ to a set of predictors $X$.

In addition to capturing interactions and approximating non-linear functions, decision trees also handle missing values in the predictors using a procedure called 'surrogate splitting'. If there are missing values on the splitting variable, a second (or surrogate) variable is selected that best approximates the original split. Individuals with missing values are then classified according to the split on the surrogate rather than the original splitting variable.

**Controlling the bias-variance tradeoff**. One of the dangers of recursive partitioning is overfitting, which refers to a model fitting the idiosyncrasies of the sample in addition to the population structure. Overfitting is a result of fitting an overly complex model, and results in high prediction error (Hastie et al., 2009). The prediction error is a function of the bias and variance of the tree, which are in turn a function of model complexity. Highly complex models have low bias but high variance, whereas low complexity models have high bias and low variance. For example, the most complex tree model is created by recursively partitioning the sample until only one observation remains in each node, achieving low bias. But the structure of this tree changes drastically from sample to sample (it has high variance), which results in high prediction error. The complexity of a decision tree can be reduced by removing unnecessary splits after a full tree is fit (pruning) or by constraining the total number of splits of the tree (Hastie et al., 2009).

**Decision tree ensembles.** A better way to reduce the prediction error of individual trees is by using decision tree ensembles. Ensembles reduce the prediction error of individual trees by aggregating the predictions from many trees. There are two primary methods to create ensembles of decision trees: *bagging* (Breiman, 1996), as used in *random forests* (Breiman, 2001; Strobl et al., 2009), and *boosting* (Freund & Schapire, 1997; Friedman, 2001). In bagging, a set of trees is fit to bootstrap samples, and the prediction of the ensemble is the average prediction across all trees. This improves prediction error by increasing the stability of the model (decreasing variance) and decreasing the influence of extreme observations. In random forests, bagging is extended by randomly selecting the set of predictors evaluated for each split in the tree, making the trees less correlated. This has the effect of further reducing the variance of the ensemble compared to bagging. Both Breiman (2001) and Strobl et al. (2009) describe random forests in more detail. We focus on a different approach for creating decision tree ensembles called boosting, which builds a tree ensemble by fitting trees that incrementally improve the predictions of the model.

### Boosting

The idea behind boosting is to iteratively create an ensemble of decision trees so that each subsequent tree focuses on observations poorly predicted by the previous trees. This is done by giving greater weight to observations that have been poorly predicted in previous trees and decreasing the weight of well-predicted observations (Freund & Schapire, 1996, 1997). Subsequent trees in the model 'boost' the performance of the overall model by selecting predictors and split points that better approximate the observations that are most poorly predicted. This procedure of iteratively fitting trees to the most poorly predicted observations was later shown to estimate an *additive model of decision trees by gradient descent* (Friedman, Hastie, & Tibshirani, 2000; Friedman 2001; Friedman, 2002). This additive model of decision trees is represented by:

$$y = F(X) = \sum_{m=1}^{M} T_m(X, \theta_m) \nu \qquad (1)$$

Where the model for an outcome variable *y* is some unknown function of the predictors $F(X)$. The goal is to approximate $F(X)$ using an additive model of $m=1,…,M$ decision trees $T_m(X, \theta_m)$ (Friedman et al., 2000). Each tree *m* has split points and splitting variables $\theta_m$. Because model (1) is a linear combination of decision trees, it retains their helpful properties: it can approximate complex, non-linear functions $F(X)$, capture interactions among the predictors, and handle missing data.

The parameters that are estimated in (1) are the splitting predictors and split points ($\theta_m$) in each tree. The parameter $v$ is called the step-size and controls how quickly the model fits the observed data. The number of trees $M$, the depth of the trees, and the step-size $v$ are meta-parameters that control the complexity of the model, and are tuned to minimize prediction error (usually by cross-validation). In the following sections, we describe how this model is estimated, tuned, and interpreted.

**Estimation of the additive model of trees by gradient descent.** A critical problem with model (1) is that the parameters, $\theta_m$, cannot be estimated in all trees simultaneously (Friedman, 2001). This is because there is no closed formula or any procedure for estimating the best possible splitting variables and split points for a single tree except by an inexhaustible computational search (Hyafil & Rivest, 1976). Estimating parameters jointly in $M$ trees is even more difficult. Because of this difficulty, the additive model of decision trees (Equation 1) is estimated by a stagewise approach called *gradient descent*. Stagewise procedures update the model one term at a time without updating the previous terms included in the model (Hastie et al., 2009). We illustrate gradient descent by first drawing a connection to multiple regression.

In multiple regression, the goal is to find the estimates of the regression weights $\boldsymbol{\beta}$ that minimize the sums of squared errors. This goal is equivalent to minimizing the squared error loss function:

$$\hat{\boldsymbol{\beta}} = \min_{\boldsymbol{\beta}} L(\boldsymbol{y}, \boldsymbol{X}\boldsymbol{\beta}) = \min_{\boldsymbol{\beta}} \sum_{i=1}^{N} (y_i - X_i \boldsymbol{\beta})^2 \qquad (2)$$

Where the parameter $\boldsymbol{\beta}$ is a vector of regression weights, $y$ is the dependent variable vector and $X$ is a matrix of predictors for $i=1, ..., N$ observations. The usual least-squares formulas that minimize the squared error loss function can be obtained by taking the first derivative (or *gradient*) of the loss function with respect to the parameters $\boldsymbol{\beta}$, setting it equal to zero, and solving for $\boldsymbol{\beta}$.

Estimating the additive model of decision trees (Equation 1) is done similarly by choosing the parameters $\theta_m$ so that the first derivative (or gradient) of the squared error loss function is minimized. In this case, the parameters are the splitting variables and split points of each tree in the model. Minimizing the loss function can be done stagewise by fitting each tree to the first derivative of the loss function, or the *gradient* (Friedman, 2001). For continuous outcome variables with squared error loss, the gradient of the squared error loss function is the vector of residuals:

$$\frac{\partial}{\partial F(X)} L(y, F(X)) = \frac{\partial}{\partial F(X)} \frac{1}{2}(y - F(X))^2 = y - F(X) \qquad (3)$$

Where $L(.)$ is the loss function, $y$ is the vector of observations, and $F(X)$ is the unknown function mapping the predictors $X$, to $y$. The derivative of the squared error loss (times the constant ½) is taken with respect to $F(X)$ at the current step $m$, so that the loss function $L$ is iteratively minimized by each tree (for details, see Friedman, 2001). Thus, estimation of the additive model of decision trees by gradient descent is equivalent to iteratively fitting decision trees to the residuals of the previous fit (Friedman, 2000, 2001).

Intuitively, this corresponds to giving greater weight to the observations that are most poorly predicted. The estimation procedure can be summarized as:

---

*Algorithm 1: Boosted Decision Trees for Continuous Outcomes Minimizing Squared Error Loss (Friedman, 2000; 2001)*

For *m* = 1, …, *M* steps (trees) do:
  1. Fit tree *m* to residuals
  2. Update residuals by subtracting the predictions of tree *m* multiplied by step-size *v*.

---

In the first step, the prediction of the model is the mean of the outcome variable, and the residuals are the deviations of the outcome around its mean.

In contrast to random forests in which trees are fully grown, individual trees are constrained to have a fixed number of splits. This is done to allow the user to directly control the degree of function approximation provided by each tree as well as the computational complexity. Empirical evidence suggests that a tree depth of 5 to 10 can capture the most important interactions (Friedman, 2001) while being relatively quick to estimate. Tuning the tree-depth often improves performance. We recommend standardizing continuous predictors prior to estimation, which makes comparisons of the relative importance of these predictors to each other more interpretable (described in more detail later).

**Re-sampling.** Re-sampling is an important improvement to the boosting procedure in which each tree is fit to a sub-sample of the observations (that is, a sample of observations drawn without replacement) at each iteration of the algorithm. Friedman (2002) showed that incorporating this stochastic sub-sampling dramatically improves the performance of the algorithm. As with random forests, this improvement results from diminishing the impact of outlying observations (Friedman, 2002). However sub-sampling is an improvement over bootstrapping because it is valid under fewer regularity conditions (Politis & Romano, 1994). The fraction of the data used to fit each tree is called the *bag fraction*, and is conventionally set at .5.

**Tuning the model by choosing the number of trees and step-size.** A critical part of successfully building this model by gradient descent is controlling the model complexity to achieve an optimal bias-variance tradeoff and low prediction error. The complexity of the model is a function of the number of trees ($M$), and the goal is to choose a minimally complex model that describes the data well. An overly complex model with many trees can fit the data too closely, resulting in high variance, whereas a model that is too simple will fail to approximate the underlying function well, resulting in high bias (Hastie et al., 2009).

There are two primary methods for choosing the best number of trees: splitting the sample into a training and test-set, or by $k$-fold cross-validation. In the first approach, the model is fit to the training set and then the number of trees is chosen to minimize prediction error on the test-set. But in this approach, the user needs to choose the fraction of the sample used for training and testing. It is often unclear how much of the sample should be used for each task. The second approach, $k$-fold *cross-validation*, provides a better estimate of the prediction error while still using the entire sample (Hastie et al., 2009). In $k$-fold cross-validation, the sample is divided into $k$ groups (called *folds,* usually 5 or 10). The model is trained on ($k$-1) of the groups, and the prediction error computed for the $k$th group. This is done for all $k$ groups so that each observation is in the test set once. The cross-validation error is the average prediction error over all $k$ groups at each step (or tree). The number of trees is then selected that minimizes the cross-validation error. Once the number of trees is chosen, the model is then re-trained on the entire sample with the selected number of trees.

The step-size ν is a meta-parameter set between 0 and 1 that indirectly affects the number of trees. It is sometimes called *shrinkage* because it shrinks the residuals at each iteration, and is sometimes called the *learning rate* because it affects how quickly the model approximates the observed data. Smaller step-sizes (e.g. .0001, .0005, or less) require many trees, but may provide better fit to the observed data (Friedman, 2001; Hastie et al., 2009). A larger step-size (e.g. .1, .5, or larger) requires fewer trees and fits the data more quickly, but can also more rapidly overfit. A typical strategy for choosing the step-size is to fix it to a small value (e.g., .001, .005, or .01) and then choose the optimal number of trees (Hastie et al., 2009). Note that the step size is a constant that cannot be factored out algebraically from (1) because trees are fit sequentially, with each tree conditional on the previous trees. Because the step size affects the residual that serves as the outcome when fitting the next tree, changing the step size results in different boosting models. The step size is fixed because computing an optimal step length based on the second derivative of the loss function (3) with respect to the parameters of tree *m* is not possible.

**Implementation**. The procedure of fitting an ensemble of decision trees by gradient descent is referred to as 'gradient boosted trees' (Elith et al., 2008) or 'gradient boosting machines' (Friedman, 2001). For univariate outcomes, the model can be estimated and tuned using the R package '*gbm*' (Ridgeway, 2013), which is an open source version of Friedman's proprietary implementation 'MART ™' and 'TreeNet™' procedures available through Salford Systems. Another open-source implementation is available in Python (Pedregosa et al., 2011).

Although outside the scope of this paper, ensembles of decision trees can also be fit to binomial, poisson, multinomial, and censored outcomes by choosing an appropriate loss function (Friedman, 2001; Ridgeway, 1999, 2013). The general procedure of fitting generalized additive models by gradient descent is known as 'boosting', and has a rich literature with many developments and extensions (see e.g. Bühlmann & Hothorn, 2007; Bühlmann & Yu, 2003, 2006; Freund & Schapire, 1996, 1997; Friedman, 2001, 2002; Friedman et al., 2000; Groll & Tutz, 2011, 2012; Ridgeway, 1999). The R package '*mboost*' (Hothorn, Buehlmann, Kneib, Schmid, & Hofner, 2015) implements boosting algorithms to estimate a wide variety of high dimensional, generalized additive models by specifying different base-procedures (e.g. splines and trees) and loss functions. For interested readers, a hands-on tutorial using *'mboost'* is also available (Hofner, Mayr, Robinzonov, & Schmid, 2014).

**Multivariate Boosting**

One goal in learning about structure with multiple outcome variables is to understand which predictors explain correlations between outcome variables. This is an important question for psychologists when multiple items are used to measure unobserved latent constructs. In factor analysis and structural equation models, the covariance between outcome variables results from a dependence on unobserved latent variables. But in a 'big data' context when a potentially a large number of predictors have been measured, it may be the case that the covariance between outcomes results from a dependence on some of the measured predictors. Examining the associations between predictors and multivariate outcomes can reveal, for example, whether predictors have similar or unique effects across the different aspects of a construct. It may also provide a different way of understanding a construct in terms of other observed or latent variables, and thus provide a basis for subsequently building large confirmatory SEMs.

We propose an extension of boosting to multivariate outcomes that selects predictors that explain covariance between pairs of outcomes. This is done by maximizing a criterion called the covariance discrepancy, denoted by $D$, at each gradient descent step. Maximizing this criterion directly corresponds to selecting predictors that explain covariance in the outcomes. To motivate this criterion, note that a single gradient descent step with squared error loss corresponds to replacing an outcome, $y^{(q)}$, with its residual at each step (Algorithm 1). In the simplest case with one dichotomous predictor and no shrinkage, the gradient descent step removes the effect of that predictor from $y^{(q)}$. If the predictor has an effect on multiple outcomes (e.g., $y^{(1,2,3)}$), the covariance between these outcomes and $y^{(q)}$ will decrease after the gradient descent step. Thus, if a predictor causes multiple outcomes to covary, there will be a discrepancy between the sample covariance matrices before and after each gradient descent step.

Formally, the covariance discrepancy $D$ is given by:

$$D_{m,q} = \left\| \hat{\Sigma}_{(m-1)} - \hat{\Sigma}_{(m,q)} \right\| \quad (4)$$

Which is the discrepancy between sample covariance matrix of the outcomes at the previous step, $\hat{\Sigma}_{(m-1)}$, and the sample covariance matrix at step $m$, $\hat{\Sigma}_{(m,q)}$, after fitting a tree to outcome $q$. The discrepancy $D_{m,q}$ quantifies the amount of covariance explained in all outcomes by the predictor(s) selected by the tree fit to $y^{(q)}$ in step $m$. At the first step, $\hat{\Sigma}_{(0)} = S$, the sample covariance matrix. $D$ corresponds to the improvement in how closely the model fits the sample covariance matrix at each step. There are many possible norms for equation (4). We employ the $L^2$ norm, which is simply the sums of squared differences between all elements of the two covariance matrices. Maximizing $D$ can be incorporated into the original boosting algorithm for squared error loss, giving Algorithm 2:

---

*Algorithm 2: Multivariate Boosting with Covariance Discrepancy Loss*

For *m* in 1, …, *M* steps (trees) do:
  1. For *q* in 1, …, *Q* outcome variables do:
      a. Fit tree $m^{(q)}$ to residuals, and compute the amount of covariance discrepancy $D_{m,q}$ (4)
  2. Choose the outcome $q^*$ corresponding to the tree that produced the maximum covariance discrepancy $D_{m,q}$ (4)
  3. Update residuals by subtracting the predictions of the tree fit to outcome $q^*$, multiplied by step-size.

---

In the first step ($m = 1$), the predictions of the model are the means of the outcome variables, and the residuals are the deviations of the outcome variables from their means. As before, the decision trees are estimated for each outcome variable by minimizing squared error loss. At each gradient descent step *m*, one tree is chosen whose selected predictors maximize the covariance discrepancy $D_{m,q}$ (4). Equivalently, the tree is chosen that maximally explains covariance in the outcome variables, or maximally improves the model implied covariance matrix. The resulting model is an ensemble of trees where the selected predictors explain covariance in the outcomes. As noted previously, we recommend standardizing continuous predictors and outcomes for interpretability and numerical stability.

      **Implementation in R**. We have developed an R package (R Core Team, 2015) called '*mvtboost*' which implements *Algorithm 2*. Our work directly extends the implementation of univariate boosted decision trees in the R package '*gbm*' (Ridgeway,

2013) such that fitting an ensemble of decision trees to a single outcome variable corresponds to using the original *'gbm'* function directly. Both *'gbm'* and *'mvtboost'* are freely available on CRAN (https://cran.r-project.org).

In the following sections, we further describe methods to tune and interpret the multivariate tree boosting model and how the *mvtboost* package can be used for these tasks in a step-by-step tutorial. For the tutorial, we use real data on the factors that predict aspects of psychological well-being in aging adults. We describe the well-being data and the research context in more detail below.

**Psychological well-being.** Identifying the factors that impact well-being in aging adults is an important step to understanding successful aging and decreasing the risk for pathological aging (Wallace et al., 2002). Previous research has identified that high resilience, coping strategies, social support from family and friends, good physical health, and the lack of stress and depression are important to successful aging (Wallace et al., 2002). In our exploratory analysis, we included these predictors as well as several additional ones — control of internal states, trait-ego resilience, and hardiness — and investigated the extent to which these predictors influenced particular aspects of well-being. Most research has focused on a well-being aggregate score, and little is known about whether the influence of these predictors varies across the different sub-scales of well-being.

A sample of 985 participants from the Notre Dame Study of Health & Well-being (Bergeman & Deboeck, 2014) filled out the surveys that were used in this analysis. The data was cross-sectional, and the age of participants ranged from 19-91 with median age 55 and with 50% of the participants falling between the ages of 43-65. More of the participants were female (58%) than male (42%).

The Psychological Well-Being Scale (Ryff & Keyes, 1995) has six sub-scales: *autonomy, environmental mastery, personal growth, positive relationships with others, purpose in life,* and *self-acceptance*. These were used as dependent variables in the analysis. Gender, age, income, and education were included as demographic predictors. The primary predictors of interest were chronic, somatic, and self-reported health, depression (separated into positive and negative indicators), perceived social control, control of internal states, sub-scales of dispositional resilience (commitment, control, and challenge), ego resilience, social support (separately for friends and family), self-reported stress (problems, emotions), and loneliness. Scale summary statistics, reliability, and missingness rates are included in Table 1, and the correlations among the well-being sub-scales are shown in Table 2. Each sub-scale is continuous, and approximately normally distributed. In total, 20 predictors were included in the analysis. All continuous predictors and the dependent variables were standardized.

*Missingness*. Well-being items were 0.3%-1% missing, with 82.6% of the participants having measurements on all items. One participant who was missing on 95% of the well-being items was removed from the analysis. Well-being sub-scales were created by averaging scores over the items ignoring missingness. The predictors had low missingness rates (1-5%), except for chronic and somatic health problems, which were not measured on the youngest cohort included in the study. Overall missingness rates were similar across gender, education, and income levels.

A modified version of the data set adding additional noise to all variables has been provided in the package '*mvtboost'* to illustrate use of the software while protecting privacy. The results reported here are from the original data, and will differ slightly from the results obtained from the data provided in the package. The original data are available upon request.

**Fitting the model using *mvtboost*.** The '*mvtboost'* package can be installed and loaded directly in an interactive R session. The well-being data set described above is included in the package, and can also be loaded into the workspace using the 'data' command. These steps are shown below:

```
install.packages("mvtboost")
library(mvtboost)
data(wellbeing)
```

To fit the model, we first assign the dependent variables to the matrix 'Y' and the predictors to the matrix 'X' using their respective column indices. After standardizing the continuous outcomes and predictors ('Ys', 'Xs' respectively) the multivariate tree boosting model can then be fit using the function mvtb:

```
res <- mvtb(Y=Ys,X=Xs)
```

Documentation for the function is available from the command ?mvtb, which describes its use in greater detail.

**Choosing the number of trees.** As with the univariate procedure, the number of trees can be chosen to minimize a test or cross-validation estimate of the prediction error. For multiple outcomes, a useful criterion is the multivariate mean-squared error:

$$MSE = \frac{1}{nQ}\sum_{i=1}^{n}(Y_i - \hat{Y}_i)^2 \qquad (5)$$

Where $Y_i$ is the vector of observations for each individual $i=1,…, n$ not used in training the model obtained for $Q$ outcome variables, and $\hat{Y}_i$ are the predicted values from the multivariate additive model of decision trees.

The default number of trees and shrinkage values for the function `mvtb` are 100 and .01 respectively. These defaults are chosen to provide a quick initial fit - further tuning is most likely necessary. For the well-being data we set the shrinkage to .005, and the maximum number of iterations to 10K. The best number of trees (2482) was chosen by five-fold cross-validation, and can be obtained using the `summary` function:

```
res <- mvtb(Y=Ys,X=Xs,n.trees=10000,shrinkage=.005,
            cv.folds=5)
summary(res)
```

**Interpreting the Multivariate Additive Model of Decision Trees**

One of the challenges of using multivariate decision tree ensembles is that the model is more difficult to interpret than a single tree. Although tree boosting can be used to build a very accurate predictive model, it is potentially more important for researchers to interpret the effects of predictors. Below, we describe approaches that have been developed to 1) identify predictors with effects on individual outcome variables, 2) identify groups of predictors that jointly influence one or more outcome variables, 3) visualize the functional form of the effect of important predictors, and 4) detect predictors with possible interaction non-linear effects.

**Predictor selection by relative influence.** The first goal in interpretation is to identify which predictors influence which outcome variables. The influence (or variable importance) of each predictor from the tree ensemble has been defined as the reduction in sums of squared error due to any split on that predictor, summed over all trees in the model (Friedman, 2001). Predictors can then be ranked by their influence or the *relative* influence, which is expressed as a percent of the total reductions in error attributed to all predictors. Predictors with large relative influence contribute more to the model than predictors with small influence. For the case of multivariate outcomes, the univariate influence is obtained for each predictor for each of the outcome variables. Summing the importance over all outcomes creates a global importance for the predictor across outcomes. To decide whether to retain a variable for further modeling, we suggest simply ranking the predictors using the influence score, and retaining a practical number of predictors higher than a given rank for any or all outcome variables. Although this suggestion is abstract, in analyses of real data it is often clear which predictors should be considered for further modeling, and may be a theoretical rather than empirical, decision.

To illustrate, consider the well-being data. The relative or raw influences of the predictors for each outcome variable can be computed using `summary` or the function `mvtb.ri`:

```
mvtb.ri(res5)
```

The results are shown in Figure 2. We see that control of internal states affects all aspects of psychological well-being except positive relationships with others. Like control of internal states, perceived stress-problems affects three aspects of well-being: self acceptance, purpose in life, and environmental mastery. Personal growth is driven by control of internal states and ego-resilience. Other patterns in the influences can be interpreted similarly, and conform to theoretical expectations (e.g. Wallace et al., 2002; Ryff & Keyes, 1995).

There are some potential issues with selecting variables based solely on the relative influence, because variable selection in trees is biased (Strobl, Boulesteix, Zeileis,

& Hothorn, 2007). This bias occurs because predictors with large variances or many categories will be selected more frequently than predictors with smaller variances or fewer categories even if the effect sizes are equal. This is because predictors with larger variances or more categories have a larger number of possible splits, and can fit more readily to idiosyncrasies in the sample. There are various approaches to correct this in random forests, and the most common is to record the difference in accuracy before and after permuting a predictor (Strobl et al., 2007). Important predictors will show large discrepancies. Like with random forests, this permutation-based procedure is available for boosted tree ensembles as well. In addition to permutation, we suggest standardizing continuous predictors to ensure that they get selected with equal priority. Standardization can also improve the interpretability of the final model by ensuring the relative influences are on the same scale.

The second issue is the case when all the predictors have zero effects in the population. In this case, predictors will still be selected into the model, and will report non-zero relative influence. As before, predictors with many categories or large variances may also be arbitrarily selected more frequently. Using the permutation procedure above can mitigate this, but the problem can be avoided altogether by assessing the fit of the model before variable selection. If the model explains little or no variance in the outcomes, there is no reason to use the model for variable selection. For the well-being data, we compute the $R^2$ for each dependent variable below. To do this, we obtain the predicted values of the model using the R function `predict`:

```
yhat <- predict(res5,newdata=Xs)
r2   <- diag(var(yhat)/var(Ys))
```

Computing the variance explained on a test-set of *n*=200 observations, the results are as follows: *autonomy* (22%), *environmental mastery* (70%), *personal growth* (42%), *positive relationships with others* (51%), *purpose in life* (57%), and *self acceptance* (50%). Other measures of model fit for multivariate outcomes can be considered as well (e.g. $\eta^2$). The model explains substantial variance in all outcomes, further substantiating our interpretation of the relative influence scores.

**Grouping predictors and outcomes by covariance explained.** In addition to selecting predictors for inclusion into a subsequent multivariate model (e.g. a multivariate regression model or SEM), it may also be informative to select the outcome variables that are associated with the set of predictors. One criterion for selecting outcome variables is to choose the outcome variables whose covariance can be explained by a function of a common set of predictors. This approach, for example, could be used to 1) identify a set of demographic predictors that similarly affect particular symptoms of a disorder, or 2) indicate to what extent covariance in sub-scales of a construct is due to effects of predictors.

The covariance explained in the outcomes by a predictor can be estimated directly by Algorithm 2. At each gradient descent step, we record the covariance discrepancy (4) *without* taking the norm (resulting in a matrix), and the predictor $X_j$ with the largest influence. Summing the raw discrepancy over all trees with each predictor approximates the *covariance explained* by each predictor. A *covariance-explained matrix* can then be organized as a *Q(Q+1)/2* x *p* table, where each element is the covariance explained in any pair of outcomes by predictor $X_j$, *j = 1, ..., p*. When the outcomes are standardized to unit variance, each element can be interpreted as the correlation explained in any pair of outcomes by predictor $X_j$. This decomposition is similar to decomposing $R^2$ in multiple regression. When the trees of the ensemble are limited to a single split and the predictors are independent, this decomposition is exact, otherwise it is approximate. The covariance-explained matrix can be used to identify groups of predictors that explain similar patterns of covariance in the outcomes. The covariance explained can be interpreted directly, or can be informative for building larger SEMs.

For the well-being data, the covariance explained matrix is obtained directly from the fitted model:

```
mvtb.covex(res5)
```

Figure 3 shows how the predictors explain correlation in pairs of sub-scales. We see that negative affect and stress problems have widespread effects on well-being. Control of internal states explains correlations across all dimensions, and is the primary explanatory predictor for autonomy. Similarly, stress, which can be detrimental to well-being, most strongly affects purpose in life and environmental mastery. Unsurprisingly, loneliness and social support from friends primarily affect positive relationships with others. Ego resilience mainly affects personal growth.

*Clustering the covariance explained matrix.* For a small number of outcomes or predictors, interpreting the covariance explained matrix is straightforward (e.g. Figure 3). But when the number of predictors or outcomes becomes large, patterns become less obvious. It can be helpful to group predictors that explain similar patterns of covariance together using clustering procedures. The *covariance explained* matrix can be clustered by first computing the distance between columns (predictors) and the rows (pairs of outcomes), respectively. Predictors that explain similar patterns of covariance in the outcomes will be closer together (have smaller distance), as will pairs of outcomes that are functions of a similar set of predictors. The resulting distance matrices computed for the rows and columns can then be used to group rows or columns by hierarchical clustering (Johnson, 1967). This corresponds to grouping the predictors that explain covariance in similar pairs of outcomes and grouping pairs of outcomes dependent on similar sets of predictors.

Clustering the covariance explained matrix can be done via the function `mvtb.cluster`. This function allows different distance metrics to be used (e.g. Euclidean, Manhattan), and different ways to cluster the distance matrices. Heatmaps or network diagrams may be useful visual aids for further interpretation. A heatmap in which the rows and columns are clustered can be obtained using the function `mvtb.heat`. The commands to cluster the covariance explained matrix for the well-being data are shown below.

```
mvtb.cluster(res5)
mvtb.heat(res5)
```

Different clustering procedures can produce alternative arrangements of the predictors and outcomes, which may suggest novel interpretations of effects. We found that grouping the rows (pairs of outcomes) without clustering produced the most interpretable solution for well-being (Figure 3) because the effects of several of the predictors concerned a single outcome variable. Examples of different clustering solutions for the covariance explained matrix are available in the package vignette. Additionally, we note that with many predictors and outcomes, it may also be helpful to cluster the matrix of relative influences. This can also be done using `mvtb.cluster` and `mvtb.heat`.

**Visualizing non-linear effects.** Another important method for interpreting the additive model of trees is to visualize predictors with non-linear effects or interactions. These plots effectively complement interpretations of relative influence by showing the direction and functional form of the effect of the predictor. Identifying non-linear effects with plots can also help prevent model misspecification if a parametric model is the final goal.

Non-linear effects can be inspected visually by plotting the fitted values of the model against individual predictors in a *partial dependence plot* (Friedman, 2001; Friedman & Meulman, 2003). In a partial dependence plot, the fitted values of the model are obtained by allowing one predictor to vary, while averaging over (or integrating out) the effects of the rest of the predictors. This plot can be extended to show the model implied effects of two variables jointly using a similar procedure. In this case, a three-dimensional perspective plot of the fitted values of the function is obtained. The fitted values are plotted jointly over a grid of the two predictors, and are obtained by averaging over the other predictors. As others have noted, these plots do not perfectly represent the effects of individual predictors, but they are still useful for interpretation (Elith, Leathwick, & Hastie, 2008; Friedman & Meulman, 2003).

Univariate and multivariate plots can be easily obtained from the *'mvtboost'* package using the base R function `plot`. For the well-being data, we plot the effect of control of internal states on personal growth (Figure 4a). From the plot we see that above-average control of internal states corresponds to larger personal growth.

```
plot(res5,predictor.no=11,response.no=3)
```

Similarly `mvtb.perspec` can be used to produce a perspective plot involving two predictors. The following code produces Figure 4b, which shows the non-additive effect of control of internal states and perceived stress problems on self-acceptance.

```
mvtb.perspec(res5,predictor.no=c(11,18),response.no=6)
```

**Detecting non-additive effects and possible interactions.** Although decision trees are models of interactions, it is difficult to detect and interpret interaction effects from a decision tree ensemble. To address this issue, we can again analyze the fitted values of the model. Following Elith et al. (2008), possible 2-way interactions can be detected by checking whether the fitted values of the approximation as a function of any pair of predictors deviates from a linear combination of the two predictors. Such departures indicate that the joint effect of the predictors is not additive, and indicate a non-linear effect or a possible interaction. A check of departures from additivity can be accomplished by computing the fitted values for any pair of predictors, over a grid of all possible levels for the two variables. For continuous predictors, 100 sample values are taken. The fitted values are then regressed onto the grid. Large residuals from this model indicate the fitted values are not a linear combination of the predictors, demonstrating non-linearity or a possible interaction. For computational simplicity with many predictors, this might be done only for pairs of important variables.

Computing the departures from additivity from the multivariate boosting model can be accomplished using the `mvtb.nonlin` function:

```
res.nl <- mvtb.nonlin(res5,Y=Ys,X=Xs)
```

This produces a large table showing the departures from additivity involving all pairs of predictors (available in a '*mvtboost*' vignette). This table can be further interpreted by plotting pairs of predictors that produced the largest departures from additivity. In the well-being example, control of internal states and stress-problems produced a high ranking departure from additivity for the dependent variable self-acceptance. This is plotted in Figure 4b. We note that this approach is primarily a heuristic for interpreting the model. A variable with a non-additive effect (e.g. a non-linear effect like control of internal states) can produce bivariate departures from additivity which are not necessarily interactions.

**Other Multivariate Boosting Approaches**

To conclude the description of multivariate tree boosting, we describe here other approaches for boosting with multivariate outcomes. For example, it is possible to use boosting to estimate a high dimensional multivariate multiple regression model by updating one component of the regression weight matrix at a time (Hothorn, Bühlmann, Kneib, Schmid, & Hofner, 2010; Lutz & Bühlmann, 2006; Obozinski, Taskar, & Jordan, 2006). It is also possible to use boosting to estimate generalized additive mixed effect models (Groll & Tutz, 2012) and non-linear time-series models (Robinzonov, Tutz, & Hothorn, 2012; Shafik & Tutz, 2009). These parametric approaches update the models one component at a time, and can use splines to transform the predictors to capture non-linear effects (Hastie & Tibshirani, 1990; Wood, 2006).

Other approaches for multivariate boosting consider the problem of classification with multiple related tasks, or *multi-task* learning. Instead of viewing multiple classification tasks as separate problems, these algorithms seek to exploit commonalities between classification tasks to improve prediction performance (Faddoul, Chidlovskii, Torre, & Gilleron, 2010). Boosting algorithms in this setting can be used for web search ranking (Chapelle et al., 2010), facial recognition from images and videos (Wang, Zhang, & Zhang, 2009), and classification of documents or e-mail (Faddoul et al., 2010). A C++ software package called *multi-boost* implements popular approaches (Benbouzid, Busa-Fekete, Casagrande, Collin, & Kégl, 2012). Our approach of estimating a model of decision trees for multivariate outcomes offers more flexibility than the parametric approaches and is more suitable for exploration with continuous, multivariate outcomes compared to these multi-task approaches.

**Variable Selection and Prediction Performance of Multivariate Tree Boosting**

We have shown how to estimate, tune and interpret the model using the R package '*mvtboost*' and the well-being data as an example. In this section we show how well the algorithm performs in comparison to other methods for effect and sample sizes that are common in psychology using simulated data. Additionally, we demonstrate the performance of the algorithm in a more traditional 'big' data context in which the number of predictors exceeds the sample size.

The performance of the algorithm is quantified in terms of variable selection performance and prediction error. The performance of multivariate boosting is compared to model-based and exploratory approaches that are often used for data exploration. The model based approaches are MANOVA and the multivariate Lasso. The exploratory approaches are multivariate classification and regression trees (De'Ath, 2002), as well as a bagged ensemble of these trees. The model based approaches are expected to perform optimally when the model is correct, but to perform poorly when the model is specified incorrectly (i.e. in the presence of non-linear effects). In these scenarios, multivariate tree boosting and (bagged) multivariate Classification and Regression Trees (CART) are expected to perform better. Below we briefly review the methods used in the simulation to select variables and build predictive models with multiple outcomes when the number of predictors is larger than the sample size.

**Approaches to Identifying Important Predictors with Multiple Outcomes**

**MANOVA.** Multivariate Analysis of Variance (MANOVA) tests for mean differences in the outcomes due to predictors (e.g. Bray & Maxwell, 1985). This is done by specifying that the multivariate response *Y* is multivariate normally distributed with mean vector **μ** and covariance matrix **Σ**. Test statistics can be formed such as Wilk's Λ, which is a ratio of determinants of the within and total sums of squares and cross product matrices. For one predictor, the distribution of Λ is known and can be used to test whether the mean vectors are significantly different between the levels of the predictor. In high dimensional settings with many predictors, predictors can be tested one at a time. This approach has been recommended for genetic association studies with multiple outcomes in statistical genetics (Ferreira & Purcell, 2009). For high dimensional contexts, CCA applied one predictor at a time is equivalent to this approach (van der Sluis, Posthuma, & Dolan, 2013).

**Multivariate Lasso.** The Lasso (Tibshirani, 1996) can be used to address the problem of estimating **β** in the multiple regression model *y* = *X***β** + **ε** when the number of predictors exceeds the sample size. It obtains estimates of **β** using least squares plus an additional penalty on the sum of the absolute sizes of the estimates $\hat{\beta}_j$, which serves to shrink some coefficients to zero. A larger penalty results in more coefficients being reduced to 0, which is useful for variable selection and reduces the variance of the estimator (Hastie et al., 2009). The trade-off however, is that all resulting estimates are biased (Tibshirani, 1996).

The Lasso for a single outcome variable can be generalized to the case of multiple outcomes by penalizing rows of **B** in the multivariate linear model $Y = XB + E$. As with the univariate Lasso, the stringency of the penalization is controlled by the meta-parameter $\lambda$. An implementation for the multivariate Lasso is available in the R package *glmnet* (Friedman, Hastie, & Tibshirani, 2010). Although the multivariate Lasso evaluates all predictors jointly even if the number of predictors exceeds the sample size, it still assumes that each outcome variable depends linearly on the predictors. To relax this assumption, spline-transformations of the predictors and product terms can be added into the model to account for non-linear effects and interactions respectively. But because multivariate tree boosting does not require specification of these effects *a priori,* we do not include these terms in the comparison. That is, we compare the methods on the basis of identical *a priori* knowledge.

**Multivariate CART.** Multivariate CART (De'Ath, 2002) is a comparable exploratory procedure to multivariate tree boosting.[1] A multivariate decision tree is fit in a similar fashion to a univariate decision tree. But instead of selecting predictors that minimize the univariate sums of squared errors, predictors are selected that minimize the sums of squared errors about the *multivariate* mean vector (De'Ath, 2002). The predictions of the multivariate tree are simply the mean of each outcome variable within each node. Because of the benefits of ensembles, a bagged version of multivariate CART was also employed, where multivariate trees were fit to 1000 bootstrap samples. Predictions from the ensemble were averaged over all trees. Splits in multivariate trees were pruned by 10-fold cross-validation. As noted previously, other methods such as saturated SEM Trees (Brandmaier et al., 2013), or multivariate conditional inference trees (Hothorn et al., 2006) are also relevant comparisons. It should be noted, however, that only this implementation was both computationally feasible for big data sets and admitted an easy to compute measure of influence.

---

[1] The implementation of multivariate CART described in (De'Ath, 2002) was not available from CRAN at the time of publication. The archived version 1.6-2 was compiled and used for the following simulations.

**Simulation Experiments**: **Variable Selection and Prediction Error**

Two simulations were carried out to quantify the performance of multivariate tree boosting for predictor selection and prediction error relative to these comparable methods. In each experiment, data was generated under a model linear in the predictors, and three models that were not linear in the predictors. For each scenario, multivariate tree boosting was compared to MANOVA testing one predictor at a time, the multivariate Lasso, and (bagged) multivariate CART. The methods were compared in terms of their variable selection performance as well as their multivariate prediction error.

**Meta-parameter selection**. For the multivariate Lasso, a value for the penalty parameter $\lambda$ (which controls the amount of penalization) was chosen using 10-fold cross-validation. For (bagged) multivariate CART, splits in each tree were only considered if they improved the fit of the tree by a fixed amount. This amount was considered as a meta-parameter and set to {.001,.0025,.005,.0075,.01,.015,.02}. Trees were then pruned by 10-fold cross-validation. For multivariate tree boosting, the maximum number of trees was fixed to 20,000, and 5 different step-size values were used

FINDING STRUCTURE IN DATA 36{.1, .01, .005, .001, .0005}. The tree depth was set to either 1 or 3, resulting in 5 x 2 = 10 conditions. Each tree was fit to a randomly selected half the sample. The best number of trees was selected by 5-fold cross-validation.

**Data generation.** Linear data was generated under the multivariate multiple regression model:

$$Y = XB + E \qquad (6)$$

Each of $p = 50$ or $p = 2000$ predictors in the matrix $X$ were independent, standard normal variables, and each error vector in the matrix $E$ was distributed standard normal. The number of outcome variables in the matrix $Y$ was 5. For the case where $p = 50$, the matrix of regression weights $B_{(50 \times 5)}$ was sparse: 15 rows each had 2 nonzero elements, so that each of these 15 predictors caused two outcomes to covary. When $p = 2000$, the matrix of regression weights was generated similarly with 100 significant predictors. The pattern of non-zero coefficients in $B$ was allowed to vary randomly across replications. The values of the coefficients were chosen to control the item-wise $R^2$. The sample size $N$ was fixed at 1000, and all observations were independent. 100 data sets were generated in this fashion.

Non-linear data was generated under the multivariate multiple regression model (6) for $p = 50$ predictors, with quadratic, cubic, and exponential transformations of the predictors. Plots of the non-linear functions are shown in Figure 5. As shown, only the exponential transformation can be well approximated by a linear model. As was the case in the linear model, the non-zero coefficients in $B$ were chosen randomly, and each predictor affected two randomly chosen outcomes. Instead of choosing the values of $B$ to control the effect size, the error variance was chosen to control a given item-wise $R^2$.

**Variable selection performance.** For each method, the following statistics were used for variable selection:

- MANOVA: The *p*-value from the *F*-test of Wilk's Λ for each predictor
- Multivariate Lasso: Penalized regression coefficients
- Multivariate CART: relative influence (the reduction in multivariate SSE attributed to splits on each predictor)
- Bagged Multivariate CART: Influence averaged over 1000 trees
- Multivariate tree boosting: The influence summed over all outcome variables.

If the test statistic was larger than a cutoff $\tau$ (smaller for MANOVA) the variable was selected. Comparing this indicator to the known true predictors produces a 2x2 table with true positives, false positives, true negatives and false negatives for a particular cutoff value (Table 3). The *true positive rate* is the ratio of true predictors that were identified by the model over the total number of true predictors. The *false positive rate* is the number of incorrectly selected predictors over the total number of predictors with truly zero effects. Ideally, a method will have a true positive rate close to 1 as well as a false positive rate close to 0. Most empirical analyses are less ideal, and choosing the cutoff $\tau$ becomes important: A liberal cutoff will lead to a high true positive rate, but also a higher false positive rate. A conservative cutoff will lead to a low false positive rate, but also a lower true positive rate.

*Measuring variable selection performance using the area under the ROC curve*. To summarize the variable selection performance independent of the threshold chosen, an ROC curve can be used (Hanley & McNeil, 1982). The curve is created by computing the true positive rate and false positive rates resulting from allowing the cutoff to take all realized values of the statistic. The true positive rates are then plotted against the false positive rates. If a procedure selects variables according to chance, the curve will be a line along the diagonal of the plot, with the true positive rate increasing along with the false positive rate. If the procedure selects variables perfectly, the true positive rate will be 1 (all true predictors selected) whereas the false positive rate is zero (no true zero predictors selected). The ROC curve can be summarized as a single number by computing the Area Under the Curve (AUC; Bradley, 1997). An AUC of .5 corresponds to chance variable selection performance, and an AUC of 1 corresponds to perfect variable selection performance. Methods can then be compared based on their AUC values, which is an indicator of performance across all possible cutoffs (Bradley, 1997).

**Prediction performance.** Prediction performance was assessed under the same data generating models used to assess variable selection performance. For each of the 100 replications, a test set was generated from the model by drawing new errors from the same distribution. The sample size for this test set was $n = 1000$. The design matrix $X$ was the same for the test set as the original sample. The multivariate mean squared error (Equation 5) was computed for each method on the $n$ new observations in the test set. The mean squared prediction error was computed directly for multivariate boosting, (bagged) multivariate CART, and the multivariate Lasso. For MANOVA, variables were first selected, and then included in a multivariate multiple regression model and the mean squared prediction error was computed from this model.

**Results**

The simulation results confirm theoretical expectations that if predictors have non-linear effects multivariate boosting performs best relative to other methods. In these cases, the methods based on the linear model are incorrectly specified. When predictors have linear effects, multivariate boosting matches the performance of MANOVA and the Lasso (showing that little power is lost) and out-performs multivariate CART.

**Variable selection performance.** The AUCs averaged over all 100 replications are shown in Figure 6. Higher AUC values indicate improved variable selection performance across all possible cutoffs. For data not well approximated by a linear model, multivariate tree boosting exceeded the performance of all methods (including bagged multivariate CART) for even very small effect sizes. When the true model was linear, or could be easily approximated by a linear model, multivariate tree boosting performed as well as the linear model methods, and better than multivariate CART or bagged multivariate CART. A similar pattern holds when selecting predictors with linear effects when $p = 2000$ (Figure 8). We note that simple multivariate CART performed very poorly in this case.

**Prediction performance.** The mean-square prediction error is shown in Figure 7. When predictors had non-linear effects, multivariate tree boosting had much lower prediction error than the Lasso or MANOVA. When the responses are truly linear functions of the predictors (or approximately linear), multivariate tree boosting performs just as well as both the Lasso and MANOVA, and better than (bagged) multivariate CART. A similar pattern holds with $p = 2000$ predictors (Figure 8), though multivariate boosting has much lower prediction error than multivariate CART and even bagged multivariate CART.

**Discussion**

Finding structure in large data collections with many predictors and outcomes is important because it can enhance content or external validity for experimental designs and provide a starting point for specifying complex parametric models in observational studies. Even with smaller data sets involving fewer variables, exploratory procedures can be helpful for detecting non-linear effects and interactions, which help to correctly specify subsequent parametric models. Finding structure requires flexible statistical methods suitable for many observed variables and little theory. Although model selection using factor models, CCA, and multivariate multiple regression models can be useful, these models make strong structural assumptions and are unwieldy with a large number of predictors. It can also be difficult or impossible to estimate parameters in a large model if the number of parameters exceeds the sample size, or to specify all possible models in order to perform a systematic model search. It can also be difficult to specify all necessary transformations of predictors to capture potential non-linear effects. Incomplete model searches likely result in ignoring predictors with important effects. Finding structure in any data set involves selecting important predictors, seeing how these predictors influence some or all outcome variables, and identifying predictors that possibly interact and have non-linear effects. To accomplish all of these things simultaneously, a highly flexible and interpretable model building approach is necessary.

A multivariate additive model of decision trees is one such approach. In this model, the outcome variables are assumed to depend on an arbitrary function of the predictors, which is then approximated by an additive model of decision trees. Decision trees provide the necessary flexibility for approximating non-linear effects and interactions without *a priori* specification, and handle missing data by surrogate splitting. Multivariate tree boosting complements existing multivariate decision tree procedures by providing methods to easily interpret the resulting ensemble. Predictors can be selected by ranking their importance in predicting each outcome, non-linear or interaction effects between pairs of predictors can be detected by testing for departures from additivity, and plots can be used to visualize these effects. Finally, we showed how to identify predictors that explain covariance in the outcomes. All of these methods contribute to a better understanding of the structure between a set of outcome variables and a set of predictors. Our model can also be used as a "black box" for prediction, if prediction rather than an investigation of variable importance is the ultimate goal.

Our simulations verified that multivariate tree boosting has better predictor selection performance and lower prediction error than other model-based and exploratory procedures when predictors have non-linear effects. The improved performance of multivariate tree boosting is dramatic even with relatively small effect sizes. Our simulations also show that when linear models are correctly specified, multivariate tree boosting performs nearly as well as methods that explicitly search for this type of effect.

Multivariate tree boosting can be used for exploratory analyses in lieu of model selection with linear models because it systematically and robustly explores existing structure in data. Both our simulations and the applied example of psychological well-being highlight the benefits of this structured exploration, which includes the discovery of predictors with non-linear effects. Multivariate tree boosting is also unambiguously interpreted as exploratory – the final results are importance scores or plots, which are suggestive and not inferential.

Multivariate tree boosting complements SEM trees and other model based recursive partitioning approaches by being maximally exploratory, highly predictive, easy to use, and still interpretable. SEM trees are best used for identifying ways to modify a known model: for example, by identifying groups with different trajectories, or groups that are not measurement invariant. SEM trees and forests also provide a 'global' measurement of variable importance in terms of the log-likelihood discrepancy between model implied covariance matrices due to splits on a predictor (Brandmaier et al., 2013). But SEM trees still make strong assumptions – all of the assumptions of SEM (even fully saturated models) must still hold within each region. In contrast, multivariate tree boosting suggests ways to build parametric models by discovering important structural features in the data while making few assumptions.

Several important practical questions remain: How large of a sample is necessary, and how many predictors can be included in the model? A sample size recommendation is difficult to make because it depends on the true model, the pattern and size of effects, and the number of variables (outcomes and predictors) under consideration. Our initial simulation results at *n = 1000,* are a useful guideline. Specific sample size limits can be investigated further using Monte Carlo simulations. More generally, statistical learning-based analyses tend to need larger samples than parametric models because the data replace the information in the structural relations specified by the parametric model.

With respect to the number of predictors, our simulation results show that boosting performs well and is computationally feasible with a very large number of predictors (i.e., 2000). Given a large enough sample and/or the presence of sufficiently large effects, investigating even larger numbers of predictors is possible given enough computation time. We note that though the number of false positives is expected to increase with a larger number of predictors, our results show that the rate of false positives can be well controlled. We caution that the variables included in the model should be selected based on their potential theoretical relevance. For questionnaire data, either items or factor scores can be included depending on the research question.

An important limitation of multivariate tree boosting that is common to all decision tree ensembles is that the estimation of individual trees and the ensemble is not optimal. An optimal approach would require an exhaustive computational search of all possible splitting variables and split points, without conditioning on a previous split or on splits in a previous tree. This limitation means that individual trees can fail to capture complex dependency structures. Further research is necessary in understanding the limits of the approximation provided by decision tree ensembles, and whether these limits are of practical importance.

There are several other limitations concerning the methods we developed for model interpretation. First, as discussed in the text, the relative influence score in this implementation is biased in favor of predictors with large variances and many categories. However, this bias can be mitigated by using alternative importance measures based on permutations of the predictors. The performance of these alternatives in the context of boosting should still be explored. Second, the departure from additivity is a heuristic, not a statistic. Further research along the lines of Mentch and Hooker (2014) is necessary to understand how well it performs in detecting interactions from boosted tree ensembles. Third, the 'covariance explained' in pairs of outcomes by predictors is only an exact decomposition for uncorrelated predictors and for decision trees with a single split. On this positive side, even these approximations can be helpful in understanding the structure in observed data. Fourth, partial dependence plots can hide heterogeneous predictor effects. Plots of independent conditional expectation may be more informative (Goldstein, Kapelner, Bleich, & Pitkin, 2015).

In addition to methodological limitations, there are shortcomings concerning the scope of the present work. For instance, our framework does not account for missingness in the outcomes. As a result, imputation by singular value decomposition, k-nearest neighbors (Troyanskaya et al., 2001), the EM algorithm (Dempster, Laird, & Rubin, 1977) or by data augmentation (Tanner & Wong, 1987) is necessary. These approaches are all likely reasonable to the extent that their assumptions about the data distribution and the missingness mechanism hold. Further research could provide a method of missing value imputation using the boosting model itself. Another limitation in scope is that our approach does not directly accommodate longitudinal data. Fitting a boosting model to factor scores in a latent growth model is one way to link an exploratory boosting model to a parametric growth model. Finally, though several representative methods of variable selection were compared in the simulation, little is known about how multivariate tree boosting influence scores compare to other variable selection methods from other models and methods not included here (for instance, variable importance in CCA and multiple regression: Grömping, 2015; Huo & Budescu, 2009; Lambert, Wildt, & Durand, 1988; Nathans, Oswald, & Nimon, 2012; Nimon, Henson, & Gates, 2010; Thompson, 2005). In general however, we can expect variable selection performance to depend on the assumptions and statistical power of the model from which the influence scores are computed.

In summary, multivariate tree boosting is a useful approach for finding structure in data for large data sets in psychology. Its flexibility makes it a compelling tool to discover and clarify important theoretical relationships that would be otherwise difficult or impossible to detect by model selection with parametric models. We hope that this work will open future developments and improvements in exploratory analyses for big data in psychology.

FINDING STRUCTURE IN DATA                                                                 51

FINDING STRUCTURE IN DATA 57Table 1: Scale sample statistics, missing rates

| Variable | M | SD | Range | α | % Missing | Scale |
|---|---|---|---|---|---|---|
| *Psychological Well-Being* | | | | | | |
| Autonomy | 2.9 | .34 | 1.5-4 | .82 | | Psychological Well-Being (Ryff & Keyes, 1995) |
| Environmental Mastery | 2.9 | .41 | 1.5-4 | .89 | | |
| Personal Growth | 3.0 | .36 | 1.1-4 | .87 | | |
| Positive Relationships | 2.9 | .43 | 1.1-4 | .89 | | |
| Purpose in Life | 2.9 | .42 | 1.4-4 | .90 | | |
| Self Acceptance | 2.8 | .47 | 1-4 | .92 | | |
| *Health* | | | | | | |
| *Chronic Health | 1.7 | 1.4 | 0-7 | .84 | 22% | Measurement of Physical Health Scale |
| *Somatic Health | 3.0 | 2 | 1-11 | .84 | 38% | |
| Self Report Health | -.02 | 4.3 | -7.9-12 | .84 | .84% | |
| *Depression* | | | | | | |
| Positive Affect | 7.4 | 3.2 | 4-16 | .67 | 3.0% | CES-D (Devins & Orme, 1985) |
| Negative Affect | 31 | 10 | 20-78 | .86 | 1.6% | |
| Perceived Social Control | 36 | 4.4 | 12-48 | .79 | 1.2% | Desired Control Measure (Reid & Ziegler, 1981) |
| Control Internal States | 51 | 6.6 | 21-71 | .91 | 2.5% | Perceived Control of Internal States Scale (Pallant, 2000) |
| *Dispositional Resilience* | | | | | | |
| Commitment | 49 | 5.8 | 17-60 | .77 | 2.0% | Hardiness (Bartone, Ursano, Wright, & Ingraham, 1989) |
| Control | 48 | 4.9 | 21-59 | .67 | 2.4% | |
| Challenge | 40 | 4.5 | 24-54 | .53 | 2.5% | |
| Ego Resilience | 42 | 6.3 | 14-56 | .84 | 1.4% | Trait Ego Reslience (Block & Kremen, 1996) |
| *Social Support* | | | | | | Perceived Social Support from Friends and Family Scale (Procidano & Heller, 1983) |
| Friends | 53 | 7.1 | 23-76 | .94 | 2.9% | |
| Family | 58 | 12 | 20-80 | .96 | 1.5% | |
| *Stress* | | | | | | Perceived Stress Scale (Cohen, Kamarck, & Mermelstein, 1983) |
| Problems | 17 | 3.4 | 7-28 | .85 | 1.7% | |
| Emotions | 15 | 3.4 | 8-28 | .85 | 1.7% | |
| Loneliness | 39 | 11 | 20-77 | .91 | 4.8% | UCLA Loneliness Scale (Russel, Peplau, & Cutrona, 1980) |

*Note:* Predictors noted by * were not measured in the youngest cohort included in the analysis (~30% of the sample).

Table 2: Correlations among psychological well-being sub-scales.

|  | Autonomy | Environmental Mastery | Personal Growth | Positive Relationships | Purpose in Life | Self Acceptance |
|---|---|---|---|---|---|---|
| Autonomy | 1 | | | | | |
| Environmental Mastery | .52 | 1 | | | | |
| Personal Growth | .46 | .57 | 1 | | | |
| Positive Relationships | .39 | .65 | .61 | 1 | | |
| Purpose in Life | .51 | .81 | .71 | .69 | 1 | |
| Self Acceptance | .54 | .82 | .61 | .67 | .86 | 1 |

FINDING STRUCTURE IN DATA                                                                 59Table 3: Confusion Matrix for variable selection.

|  | Effect of Predictor | |
| --- | --- | --- |
| Selected by Method | $\beta_j > 0$ | $\beta_j = 0$ |
| $\hat{I}_j > \tau$ | TP | FP |
| $\hat{I}_j < \tau$ | FN | TN |

*Note:* Columns: Predictor $j$ has a true effect in the population if $\beta_j$ is greater than zero or equal to zero. Rows: Predictor $j$ is labeled as having an effect in the population if statistic $\hat{I}_j$ is greater or less than a cutoff $\tau$. In the simulation, the statistic is the influence measure from the decision tree methods (boosting, multivariate CART, and bagged multivariate CART), *p*-value from MANOVA, or the estimated regression coefficient from the Lasso. For MANOVA, the rows of this matrix are reversed because variables are selected when $p < \tau$, where $\tau = \alpha$. TP=true positive, FP = false positive, FN = false negative, TN = true negative.

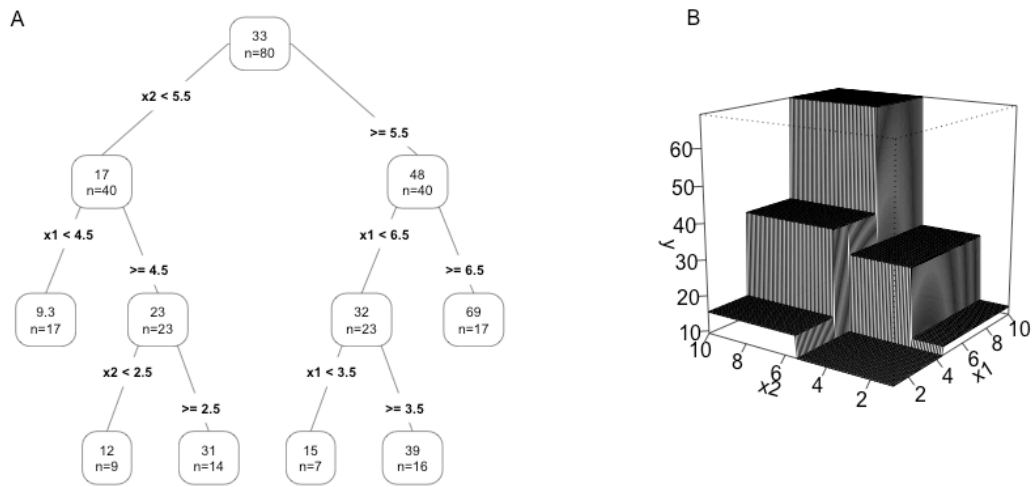

*Figure 1:* Representations of decision trees. Representation of a decision tree as a tree diagram (Panel A) and as a surface in three dimensions (Panel B) for two predictors. In the decision tree (Panel A) the means (top) and sample sizes (n = …) within each node are shown, and the split is shown in each branch. (Panel B) illustrates that decision trees are a piecewise-constant approximation of non-linear and interaction effects: each split divides the predictor space into rectangular regions, and the prediction of the tree is the mean of the response variable in each region. Plots following Hastie et al. (2009), Elith (2008), and many others.

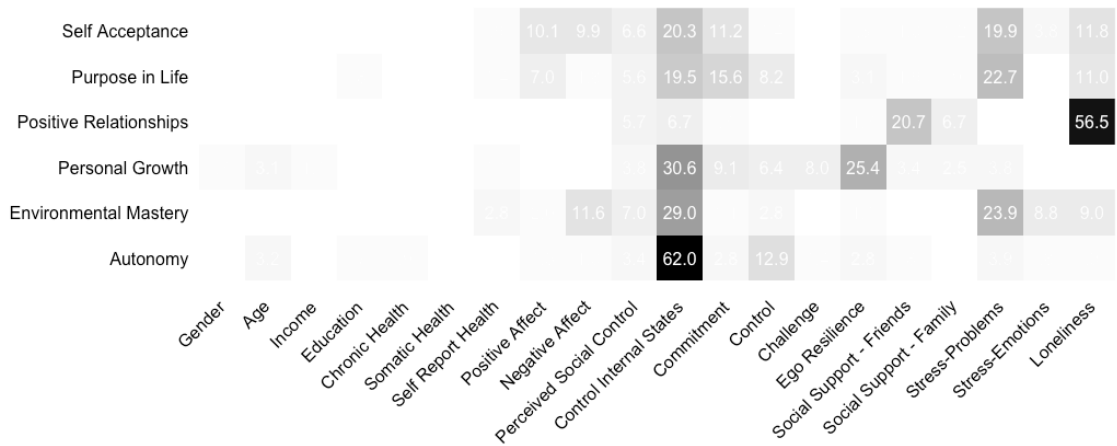

*Figure 2*. Relative influences from multivariate tree boosting. The relative influences are the sum of squared reductions in error attributable to splits on that predictor, and are reported as percent of the total reduction sums of squares for each well-being sub-scale. We see that control of internal states is important across well-being sub-scales. Control of internal states also has the single biggest contribution of any predictor, with a large effect autonomy. Loneliness, stress – problems, ego resilience, and social support from friends are also important to aspects well-being.

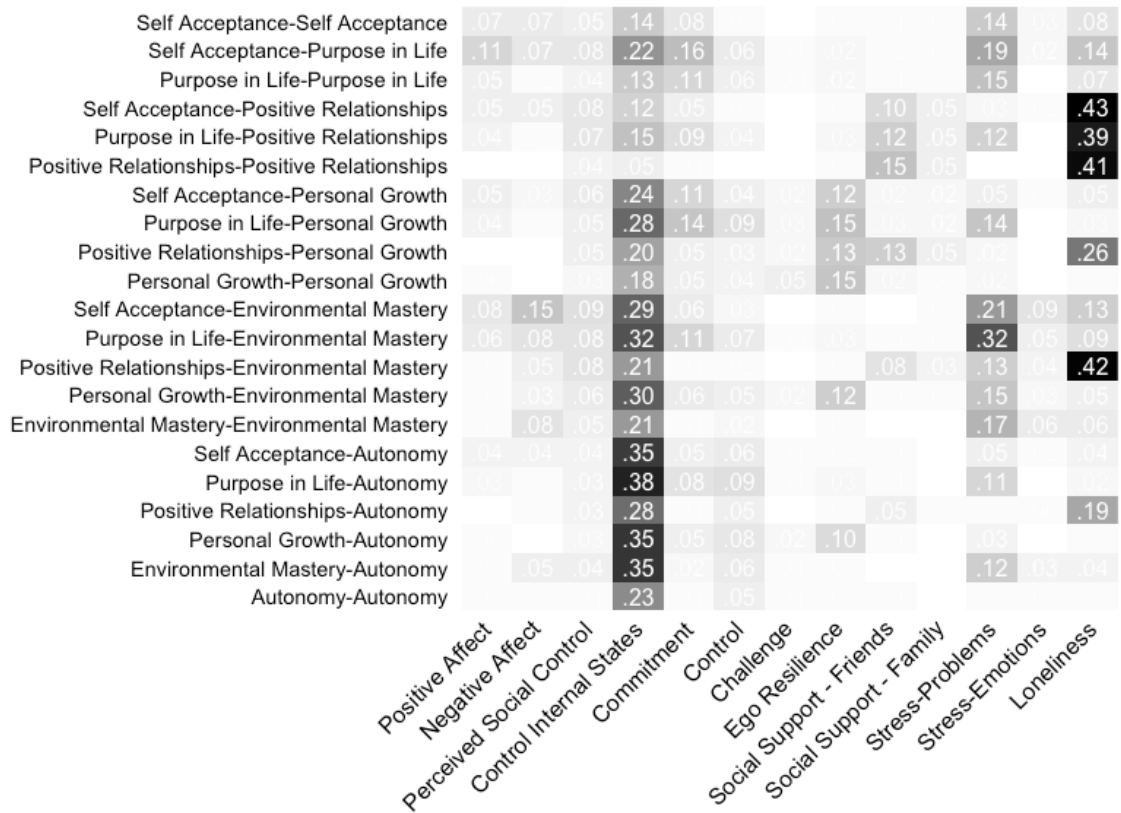

*Figure 3:* Correlation explained in pairs of sub-scales by a subset of the predictors. The amount each predictor (columns) explained variance or covariance in any pair of sub-scales (rows). Predictors with very small or zero effects included chronic/somatic health, income, education, gender, and age have been removed. Control of internal states and explains correlation across almost all dimensions, and is the primary explanatory predictor for autonomy. Stress problems primary affects purpose in life and environmental mastery. Loneliness mainly affects positive relationships and the correlation between positive relationships and other factors. Ego-resilience mainly affects personal growth, and social support from friends is associated with positive relationships.

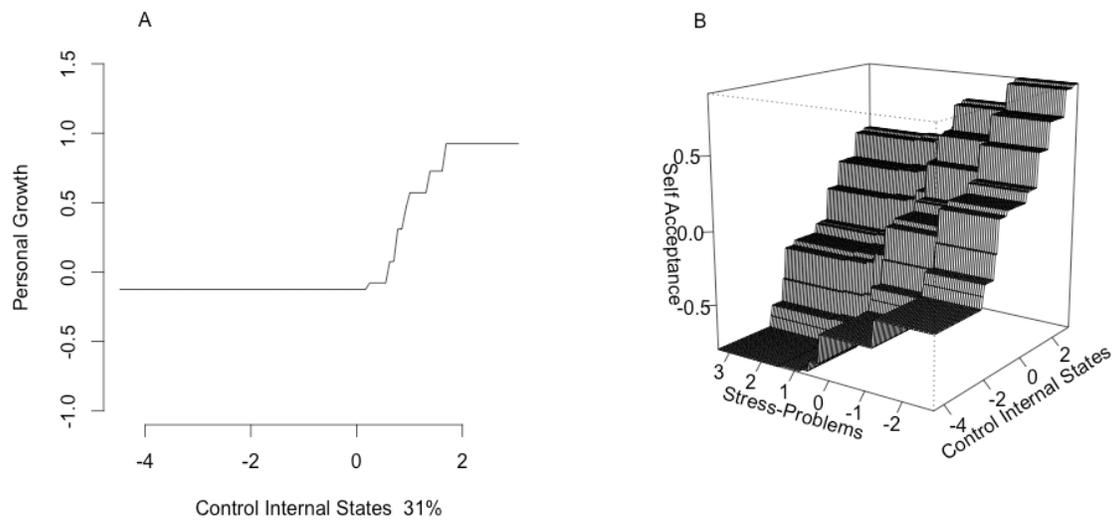

*Figure 4.* Model implied effects of control of internal states. Because the predictors and outcome were standardized, unit changes on the *x*, *y* and *z* axes correspond to standard deviation changes in the predictors and outcomes. Panel A shows the predicted values of personal growth as a function of control of internal states (with % relative influence). Control of internal states shows a strong non-linear effect – above average control is associated with larger personal growth. Panel B shows the model predicted values for self acceptance as a function of stress problems and control of internal states, indicating a possible multiplicative effect rather than simply an additive one.

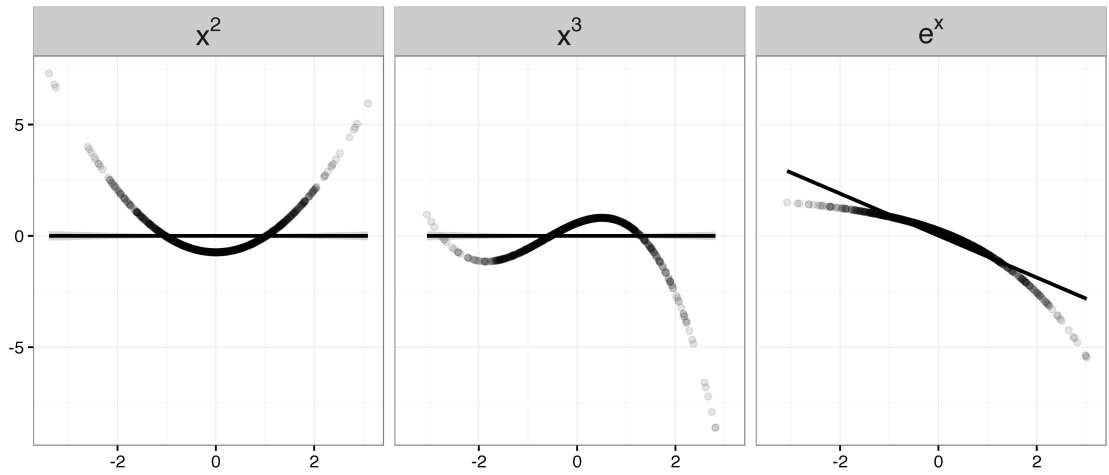

*Figure 5*. Simulated non-linear effects. Simulated non-linear effects for a single predictor: $x^2$, $x^3$, $e^x$ for 1000 samples. Only $e^x$ is well approximated by a linear model.

FINDING STRUCTURE IN DATA 65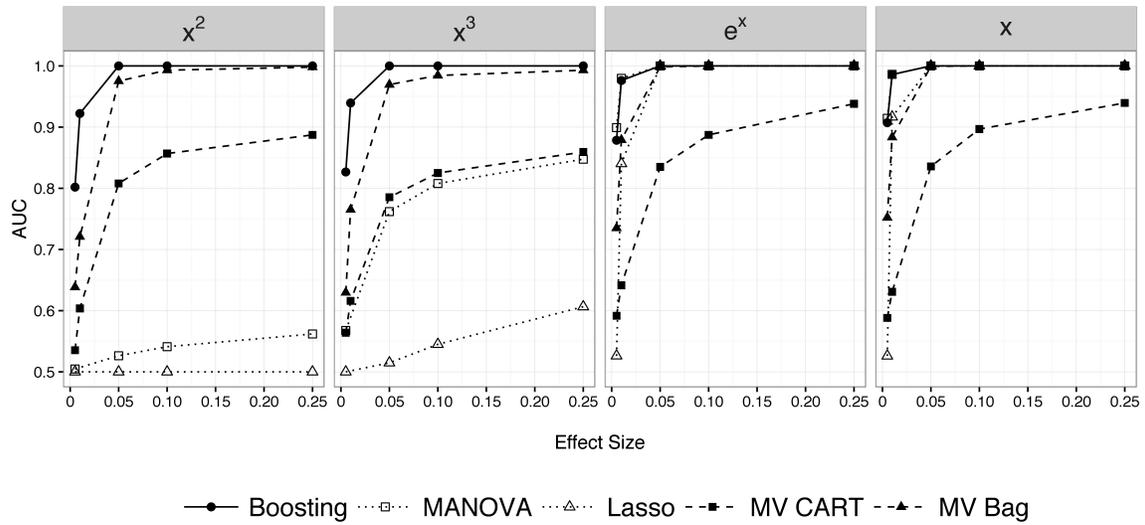

*Figure 6*. Variable selection performance for simulated data. Variable selection performance of multivariate boosting compared to MANOVA, the Lasso, multivariate CART, and bagged multivariate CART. The performance (higher is better) is shown for a given non-linear effect ($e^x$, $x^2$, $x^3$, $x$ is the linear model) for a range of effect sizes. Higher AUCs indicate better performance: an AUC of 1 is perfect and an AUC of .5 corresponds to variable selection no better than chance. For effects that are not linear ($x^2$, $x^3$), boosting dominates all other methods followed by bagged multivariate CART. For $e^x$ and $x$ (linear effect), boosting does not perform significantly worse than MANOVA or the Lasso.

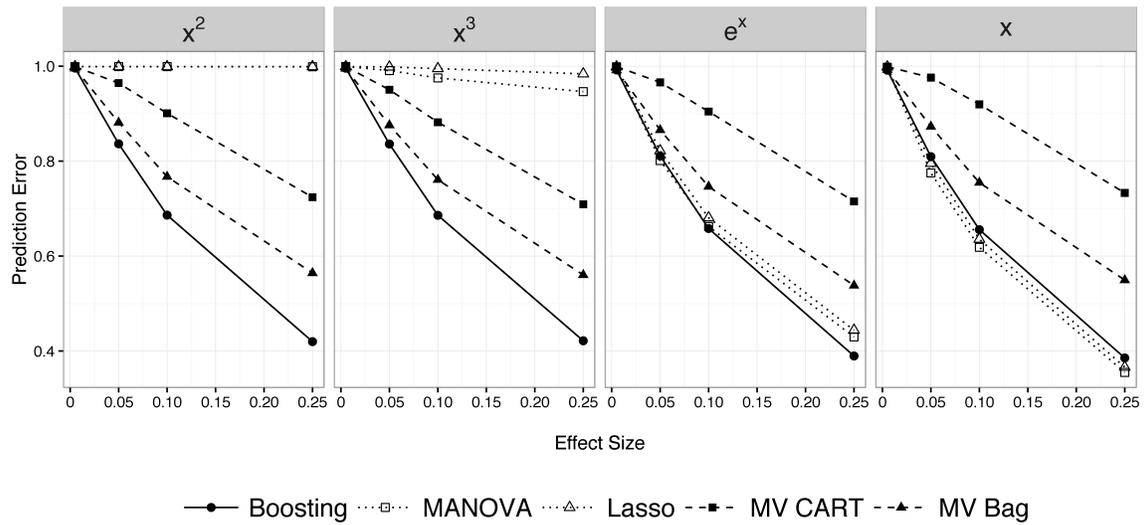

*Figure 7.* Prediction performance for simulated data. Prediction performance of multivariate boosting, MANOVA, the Lasso, multivariate CART, and bagged multivariate CART. Lower prediction error is better. The performance is shown for a given non-linear effect ($e^x, x^2, x^3$, $x$ is the linear model) for a range of effect sizes. Multivariate boosting has lower prediction error than bagged multivariate CART, MANOVA or the Lasso for conditions with non-linear effects ($x^2, x^3$). It has comparable performance under effects that are linear ($x$) or approximated well by a linear model ($e^x$).

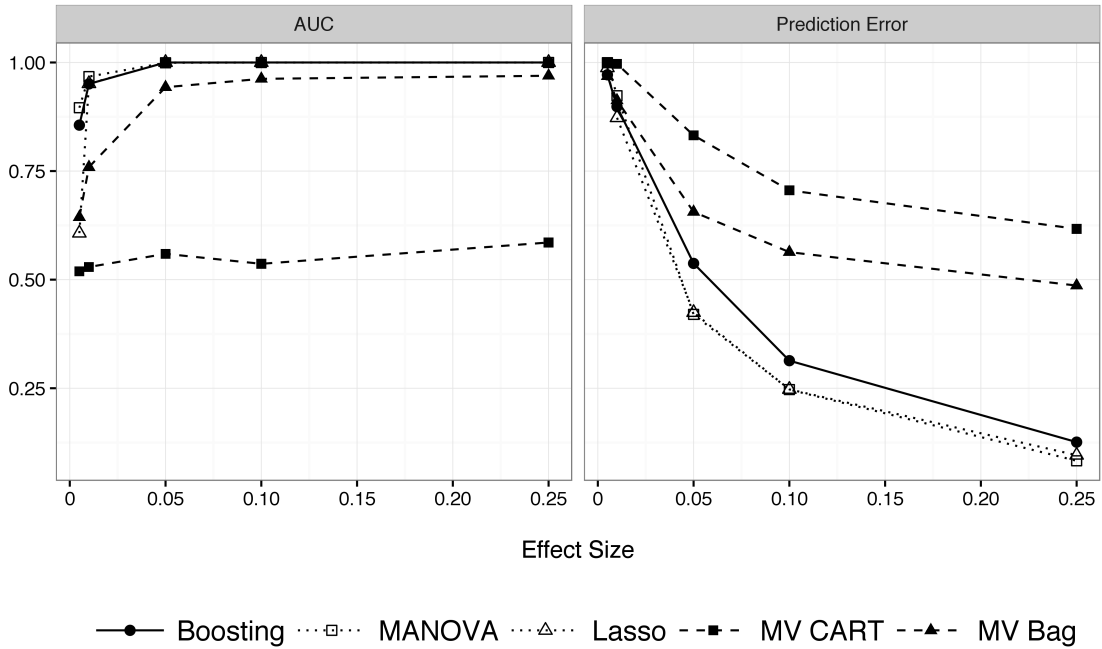

*Figure 8.* Prediction and variable selection performance for simulated data with *p* = 2000 > *n*. Prediction performance of multivariate boosting, MANOVA, the Lasso, multivariate CART, and bagged multivariate CART when *p* > *n*. The performance is shown when predictors have linear effects only. Multivariate boosting does not perform worse than MANOVA or the Lasso, and performs much better than multivariate CART and bagged multivariate CART.